\documentclass{article}

\PassOptionsToPackage{numbers, sort&compress}{natbib}

\usepackage[preprint]{neurips_2023}

\usepackage[utf8]{inputenc} 
\usepackage[T1]{fontenc}    
\usepackage[colorlinks]{hyperref}       
\usepackage{url}            
\usepackage{booktabs}       
\usepackage{amsfonts}       
\usepackage{nicefrac}       
\usepackage{microtype}      
\usepackage{xcolor}         
\usepackage{colortbl}
\usepackage[normalem]{ulem}
\definecolor{ai21color}{RGB}{255,75,137}

\usepackage{multicol}
\usepackage{xspace}
\usepackage{graphicx}
\usepackage{subcaption}
\usepackage{todonotes}
\usepackage{multirow}
\usepackage{makecell}
\usepackage{enumitem}
\usepackage{stfloats}
\fnbelowfloat
\usepackage[compact]{titlesec}

\def\jamba{Jamba\xspace}
\def\jambamini{Jamba-1.5-Mini\xspace}
\def\jambalarge{Jamba-1.5-Large\xspace}
\def\newquant{ExpertsInt8\xspace}

\title{\jamba-1.5:\\ Hybrid Transformer-Mamba Models at Scale
}

\author{%
Jamba Team
}

\begin{document}

\maketitle

\begin{center}
\vspace{-20pt}
\centering
\includegraphics[width=0.30\linewidth,keepaspectratio]{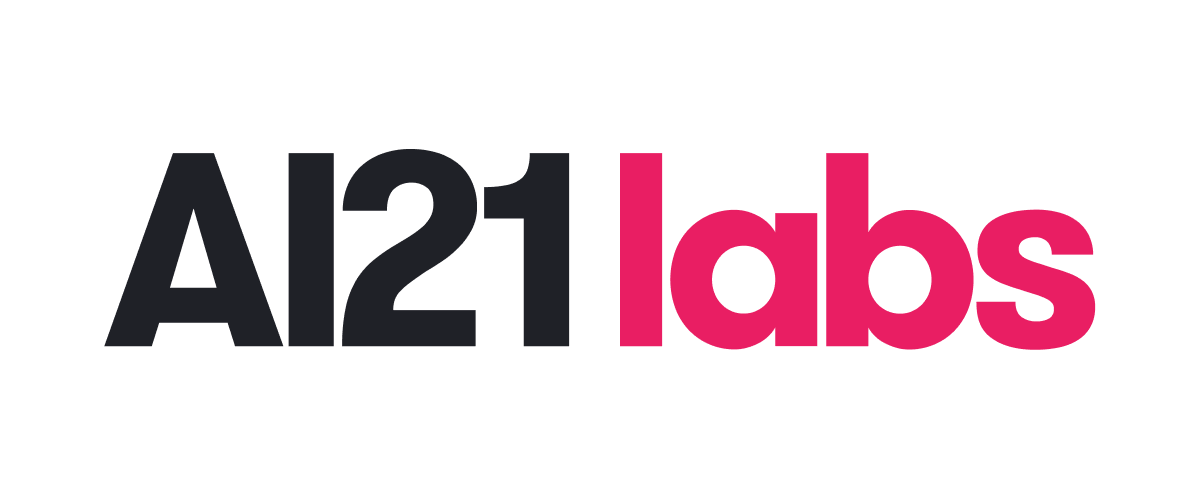}
\end{center}

\vspace{10pt}

\begin{abstract}
We present \jamba-1.5, new instruction-tuned large language models based on our \jamba architecture. \jamba is a hybrid Transformer-Mamba mixture of experts architecture, providing high throughput and low memory usage across context lengths, while retaining the same or better quality as Transformer models. We release two model sizes: \jambalarge, with 94B active parameters, and \jambamini, with 12B active parameters. Both models are fine-tuned for a variety of conversational and instruction-following capabilties, and have an effective context length of 256K tokens, the largest amongst  open-weight models. To support cost-effective inference, we introduce \newquant, a novel quantization technique that allows fitting \jambalarge on a machine with 8 80GB GPUs when processing 256K-token contexts without loss of quality. When evaluated on a battery of academic and chatbot benchmarks, \jamba models achieve excellent results while providing high throughput and outperforming other open-weight models on long-context benchmarks. The model weights for both sizes are publicly available under the \jamba Open Model License and we release \newquant as open source.   \\\\
 \textbf{Models:}  \url{https://huggingface.co/ai21labs}

\end{abstract}

\section{Introduction}
\label{sec:intro}

This paper introduces \jamba-1.5, two new large language models based on our \jamba architecture \cite{Lieber2024JambaAH}, which are available for public use. 
\jambamini is an updated and instruction-tuned version of our earlier \jamba release \cite{Lieber2024JambaAH}. 
Like its smaller sibling, \jambalarge is a hybrid architecture that mixes Transformer \cite{vaswani2017attention} and Mamba  \cite{gu2023mamba} layers, with a mixture-of-experts (MoE) module \cite{shazeer2016outrageously,fedus2022switch}. 
Since the introduction of \jamba, similar efforts have confirmed the benefits of combining Transformer and state-space-models at a scale of up to 8B parameters \cite{Dao2024TransformersAS,Waleffe2024AnES}.  \jambalarge demonstrates the benefits of this architecture at a much larger scale. 
It has 94B active parameters, out of a total of 398B parameters. Even at this large size, the model can fit on a single machine with 8 80GB GPUs when processing a context of 256K tokens, thanks to the efficiency of the \jamba architecture in addition to a novel quantization technique we have developed, as described in Section \ref{sec:quant}. 

Both \jambamini and \jambalarge  are instruction-tuned models, having undergone post-training to provide them with various capabilities. Our evaluations across a wide range of benchmarks show that they perform comparably to models at their size, while offering the efficiency benefits of the \jamba architecture. In particular, \jamba-1.5 models shine at long-context evaluations, making them the only models with an effective length of 256K on the RULER benchmark, while offering 10x reduction in KV cache memory as well as superior throughput and latency.

We make the \jamba-1.5 models available under the Jamba Open Model License:\\
\url{https://www.ai21.com/licenses/jamba-open-model-license}.\\
The models are publicly available:\\
\textbf{\jambamini:} \url{https://huggingface.co/ai21labs/AI21-Jamba-1.5-Mini}\\
\textbf{\jambalarge:} \url{https://huggingface.co/ai21labs/AI21-Jamba-1.5-Large}

\section{Model Architecture} \label{sec:arch}

\jambalarge is based on \jamba \cite{Lieber2024JambaAH}, our hybrid decoder architecture that mixes Transformer layers \cite{vaswani2017attention} with Mamba layers \cite{gu2023mamba}, a state-space model (SSM) \cite{gu2021combining,gu2021efficiently}, in addition to a mixture-of-experts (MoE) module \cite{shazeer2016outrageously,fedus2022switch}. See \cite{Lieber2024JambaAH} for a detailed description of this architecture. 

During our work on \jamba \cite{Lieber2024JambaAH}, we found that the combination of Transformer, Mamba, and MoE elements facilitates balancing desiderata of throughput, memory usage, and quality. \jambalarge demonstrates this flexibility at a larger scale. 

\jambalarge follows the same \jamba structure but with a larger capacity. It has 94B active parameters and 398B total parameters. It has $9$ blocks, with each block having the following specs:
\begin{itemize}
    \item $l=8$ layers in each block.
    \item $a:m = 1:7$ ratio of attention-to-Mamba layers. This ratio was found optimal in our work on \jamba \cite{Lieber2024JambaAH} and similar ratios was also confirmed as successful in follow-up work \cite{Dao2024TransformersAS,Waleffe2024AnES}.
    \item MoE is used instead of a single MLP every $e = 2$ layers. There are $n=16$ experts, and we select the top $K=2$ at each token.
    \item The hidden state dimensionality is $8192$. 
    \item The number of attention query heads is $64$ and the number of KV heads is $8$. 
\end{itemize}

Table~\ref{table:params-cache} compares the \jamba-1.5 models to publicly available models of similar sizes. \jambamini has a similar number of active parameters as Mixtral 8x7B, while \jambalarge's active parameter count is between LLaMA-3.1-70B and Mistral-Large-2. 
At the same time, both our \jamba models have a much smaller KV cache memory usage (at 256K tokens) compared to all other models, with roughly an order of magnitude reduction compared to their respective counterparts.

With these settings, and our specialized quantization (Section \ref{sec:quant}), \jambalarge can be served on a single machine with 8 80GB GPUs with context lengths up to 256K tokens. 

\begin{table}[h]
\centering
\begin{tabular}{l r r c}
    \toprule
    & \multicolumn{1}{c}{Available params} & \multicolumn{1}{c}{Active params} & \multicolumn{1}{c}{KV cache (256K context, 16bit)} \\ 
    \midrule     
    Mistral & 7.2B & 7.2B & 32GB \\ 
    Mixtral 8x7B & 46.7B & 12.9B & 32GB \\ 
    LLaMA-3.1 8B & 8B & 8B & 32GB \\ 
    Mixtral 8x22B & 141B & 39B & 56GB \\ 
    Mistral-Large-2 & 123B & 123B & 88GB \\ 
    LLaMA-3.1 70B & 70B & 70B & 80GB \\ 
    LLaMA-3.1 405B & 405B & 405B & 252GB \\ 
    \midrule 
    \textcolor{ai21color}{\bf \jambamini} & 52B & 12B & 4GB \\
    \textcolor{ai21color}{\bf \jambalarge} & 398B & 94B & 9GB \\ 
    \bottomrule
\end{tabular}
\vspace{1pt}
\caption{Comparison of \jambamini, \jambalarge and recent open models in terms of total available parameters, active parameters, and KV cache memory on long contexts. \jambamini and \jambalarge provide substantial reductions in the KV cache memory requirements.}
\label{table:params-cache}
\end{table}

\bigskip \bigskip  

\pagebreak 

\begin{figure}[h]
    \centering
    \begin{subfigure}{\textwidth}
    \includegraphics[width=\linewidth]{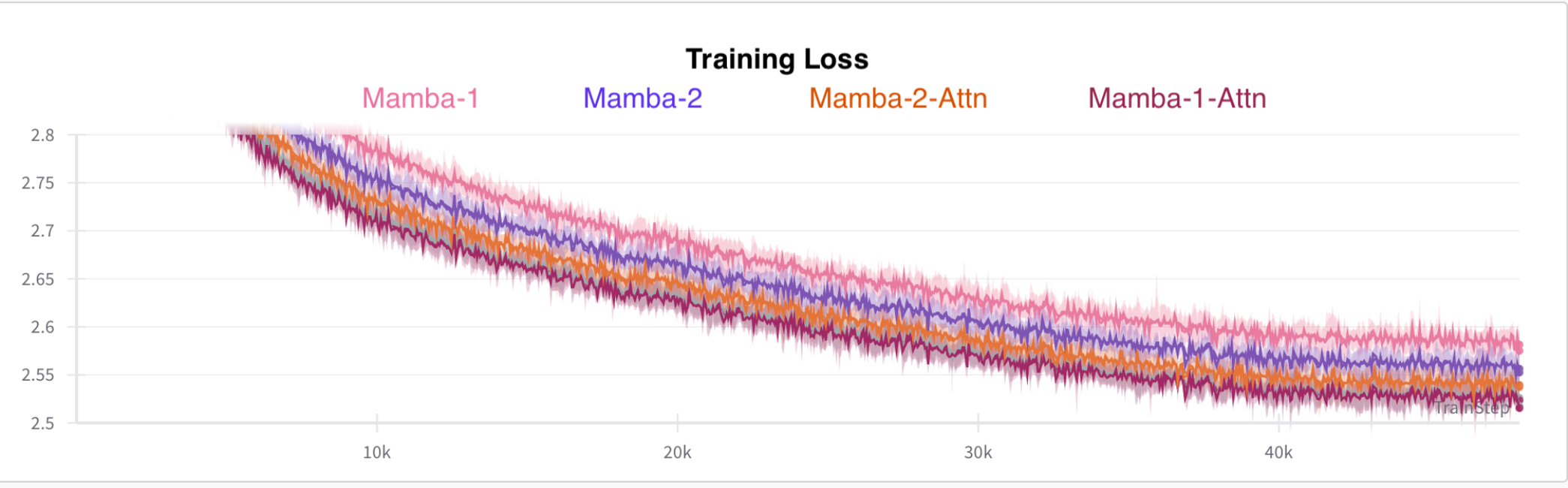}
    \caption{350M.}
    \label{fig:mamba1-2-350}
    \end{subfigure}\\
    \begin{subfigure}{\textwidth}
    \includegraphics[width=\linewidth]{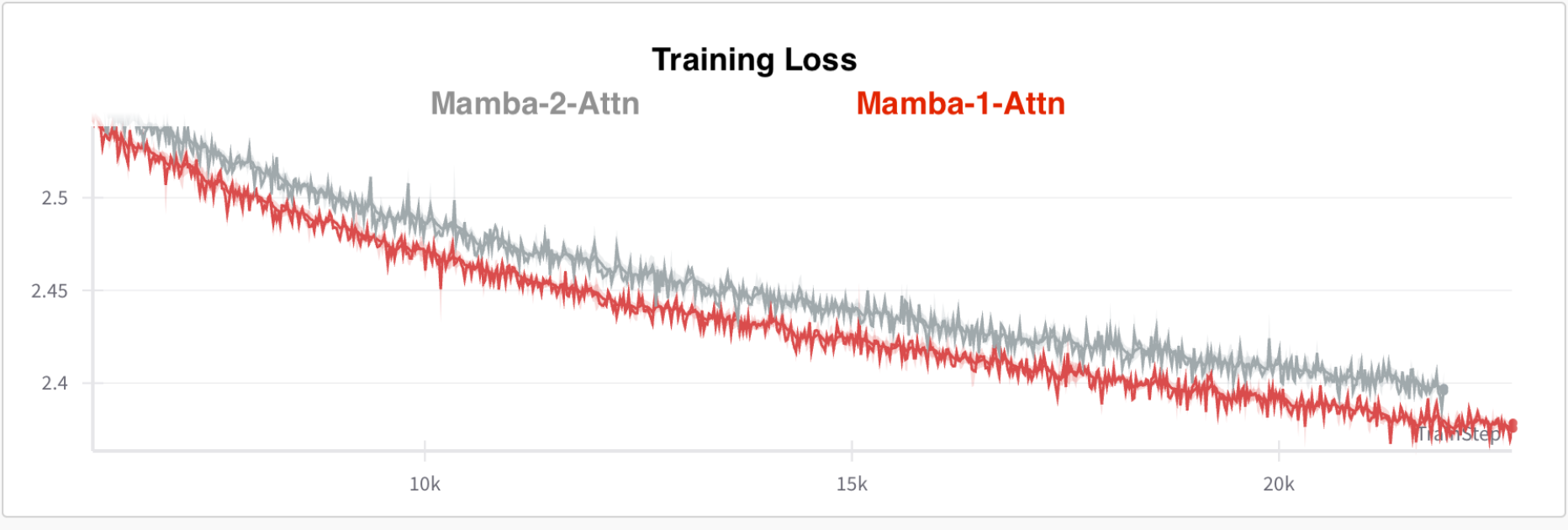}
    \caption{1.3B.}
    \label{fig:mamba1-2-1.3}
    \end{subfigure}
    \caption{Comparison of Mamba-1, Mamba-2, Mamba-1-Attention, and Mamba-2-Attention on models trained for 100B tokens. While Mamba-2 outperforms Mamba-1 without attention, the hybrid Mamba-1-Attention performs better.}
    \label{fig:mamba1-2}
    \vspace{-10pt}
\end{figure}

For this release, we experimented also with Mamba-2 \cite{Dao2024TransformersAS}, a faster and improved version of Mamba, which was reported to outperform Mamba and Transformers separately. However, as Figure \ref{fig:mamba1-2} shows, we found that in a hybrid architecture, the Mamba-1-Attention combination works better than Mamba-2-Attention, so we use Mamba-1 in \jambalarge. (We also found the hybrid architecture to outperform pure Mamba-2.)
We hypothesize this is because some  of the advantages of Mamba-2 over Mamba-1, in particular the ability to use a much larger state size, are less significant when we have full attention layers interleaved between the Mamba layers, as they can pool information from the entire context. 

\section{Serving Considerations and Improvements}
We share a few insights and improvements we have introduced to allow for efficient serving of \jamba models at a large scale.

\subsection{\newquant Quantization} \label{sec:quant}
To support efficient serving of \jambalarge, we developed a new quantization technique, which we dub \newquant. We observe that over 85\% of the model weights are in the MoE layers, and over 90\% are in MoE or MLP layers. We wish to quantize these weights while still enjoying the benefits of fast BF16 kernels. To do so, we quantize the MoE and MLP weights to INT8, save them in INT8, and dequnatize them back to BF16 before the actual computation. Importantly, the dequantization step happens directly inside the \texttt{fused\_moe} kernel in vLLM \cite{kwon2023efficient}.
In this way, the dequantization process adds negligible overhead, and even leads to improved latency over BF16.\footnote{We attribute this to the the kernel operating on relatively small blocks of weights and activations, which it moves from GPU HBM to SRAM prior to  performing the computations. In our implementation, the weights move from HBM to SRAM when they are in int8, so it takes less time as their memory footprint is cut by half.} We  have contributed our modified \texttt{fused\_moe} kernel to vLLM.\footnote{Pull request here: \url{https://github.com/vllm-project/vllm/pull/7415}}  

Our \newquant method has several advantages. First, it is fast; quantization only takes a few seconds at model loading. Second,  unlike most other techniques in vLLM, it does not rely on  calibration, which can take hours or days and can be unstable. Third,  we can still use BF16 to hold large activations. Fourth, it is available to use on A100 GPUs, unlike FP8, which is only available on H100. Finally, our quantization matches FP8 in latency, while surpassing other quantization techniques, without a loss in quality.

\begin{figure}[h]
    \centering
    \begin{subfigure}{.33\textwidth}
    \includegraphics[width=\linewidth]{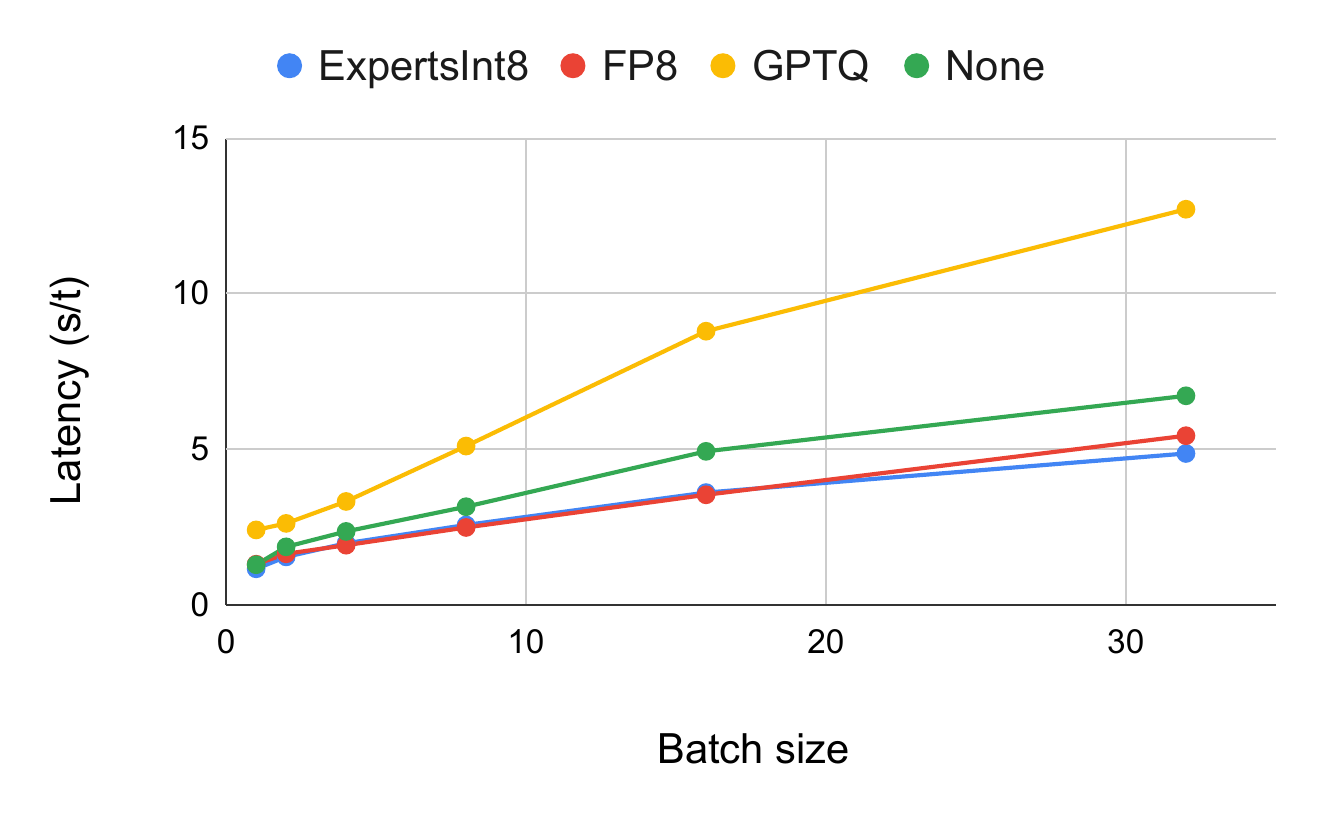}
    \caption{\jamba, 2xH100}
    \label{fig:quant-jamba-2h100}
    \end{subfigure}%
    \begin{subfigure}{.33\textwidth}
    \includegraphics[width=\linewidth]{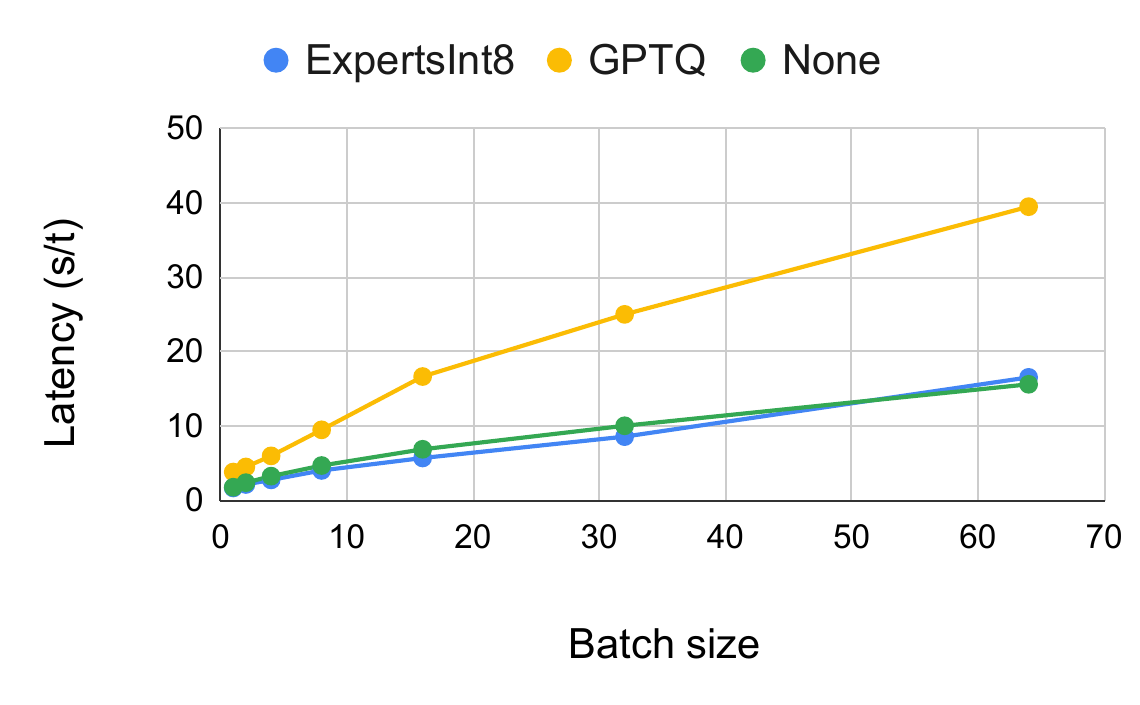}
    \caption{\jamba, 2xA100}
    \label{fig:quant-jamba-2a100}
    \end{subfigure}%
    \begin{subfigure}{.33\textwidth}
    \includegraphics[width=\linewidth]{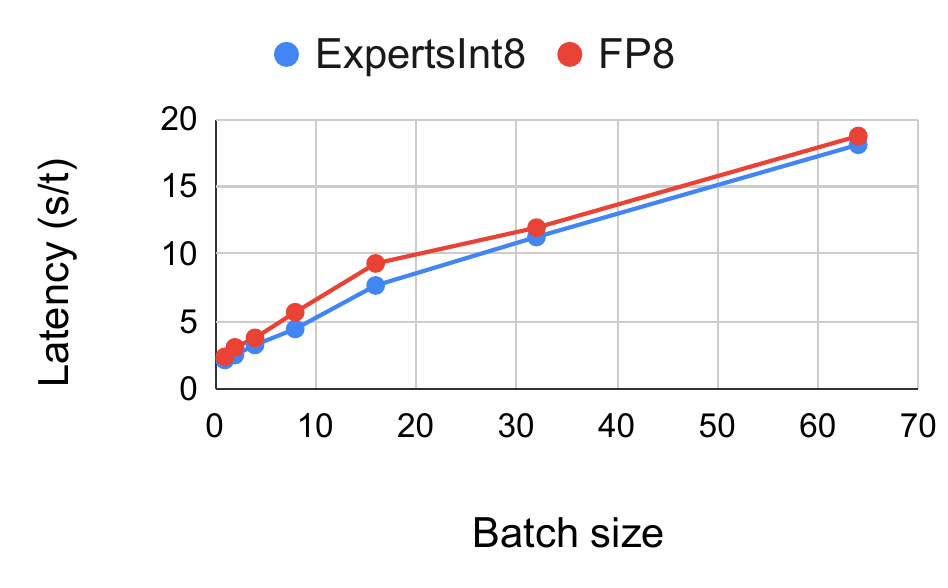}
    \caption{\jambalarge, 8xH100}
    \label{fig:quant-jamba-8h100}
    \end{subfigure}\\   
    \begin{subfigure}{.33\textwidth}
    \includegraphics[width=\linewidth]{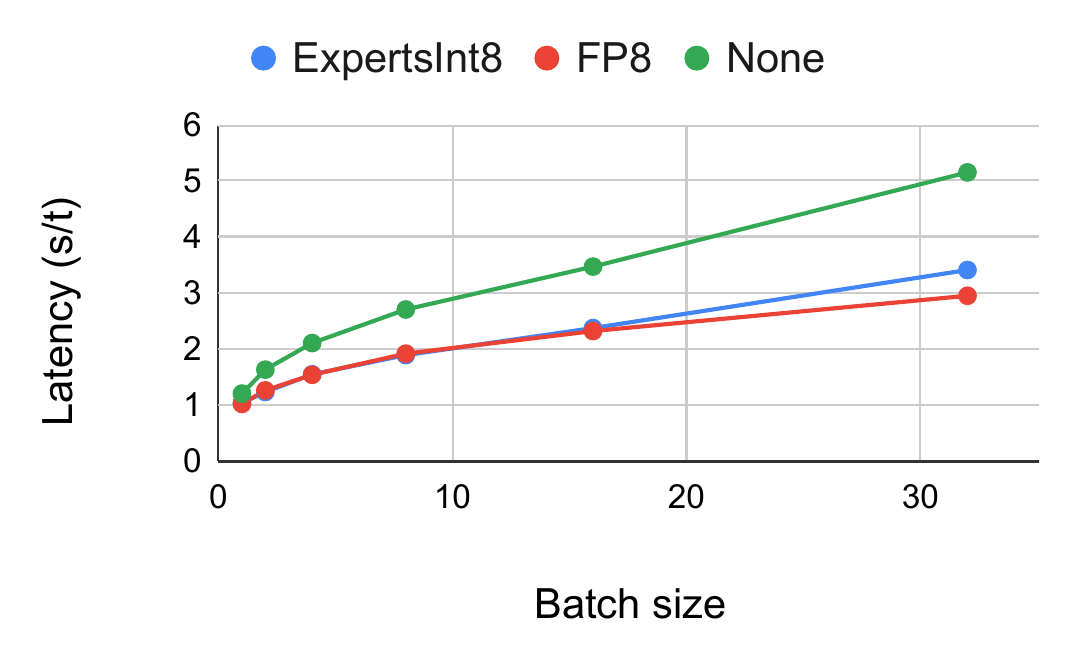}
    \caption{Mixtral-8x7B, 2xH100}
    \label{fig:quant-mixtral-8x7b-2h100}
    \end{subfigure}%
    \begin{subfigure}{.33\textwidth}
    \includegraphics[width=\linewidth]{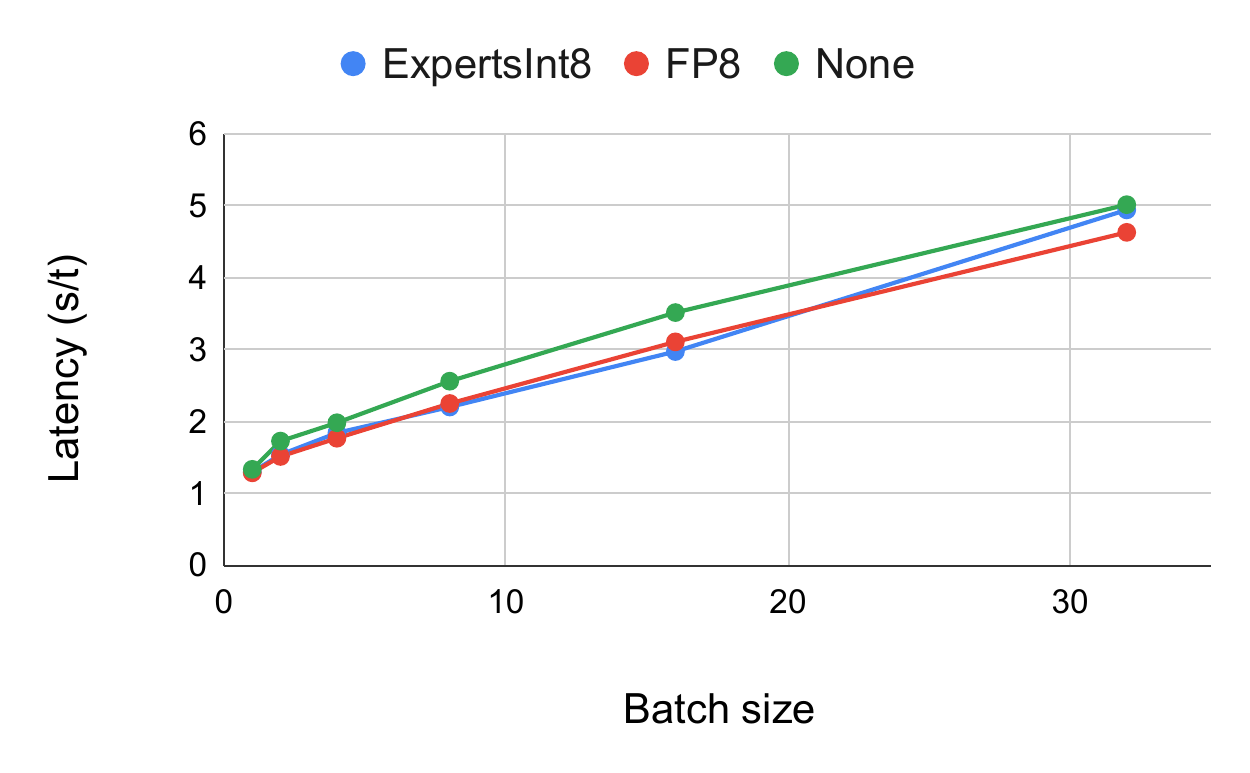}
    \caption{Mixtral-8x22B, 8xH100}
    \label{fig:quant-mixtral-8x22b-8h100}
    \end{subfigure}\\       
    \caption{Comparison of different quantization techniques, showing end-to-end latency with 1024-token context and 128-token decoding. \newquant performs similar to FP8, while being fast and simple to apply and still allowing BF16 activations, as well as applicable to A100 GPUs, where FP8 is unavailable.   }
    \label{fig:quant}    
\end{figure}

Figure \ref{fig:quant} compares the latency with different quantization techniques using \jambamini, \jambalarge, and two Mixtral models (8x78B and 8x22B). On H100 GPUs, \newquant matches the latency of FP8. On A100, where FP8 is unavailable, \newquant is an attractive technique, outperforming GPTQ \cite{frantar2023optq} by a large margin. Together with the advtanages of \newquant explained above, this makes it an attractive quantization technique for serving large MoE models. 

\subsection{Activation Loss}
During pre-training, we found that certain activations, namely outputs of specific experts as well as the the output of the last Mamba layers, were gradually increasing in magnitude for certain input tokens, eventually reaching values as high as $4\times 10^{6}$. 
Although we did not find this to hurt the pre-training itself, which was done in BF16 precision, the magnitude of the activations could cause numerical issues during inference as some quantization libraries support only FP16 precision for activations, which has a maximum range of 64K.

To alleviate these concerns, we added an ``Activation Loss'' term, proportional to the mean-square of activations in the forward pass, with a configurable $\alpha$ factor, which penalizes larger activation values.
We found via experimentation that this auxilary loss has no affect on the training even with $\alpha$ values up to at least $10^{-3}$. For \jambalarge, we used $\alpha = 10^{-5}$ which was enough to reduce the activations to an acceptable range (2K-3K max). Moreover, adding this auxilary loss reduced the activations almost instantly, allowing it to be added only towards the end of the training without any affect on training speed and quality.

To validate this approach, we ran our full evaluation suite on the model using FP16 activations and obtained the same results as the BF16 evaluations without any nans/overflows.

\bigskip 
\pagebreak

\section{Throughput and Latency Analysis}

\begin{figure}[h]
    \centering
    \begin{subfigure}{.49\textwidth}
    \includegraphics[width=\linewidth]{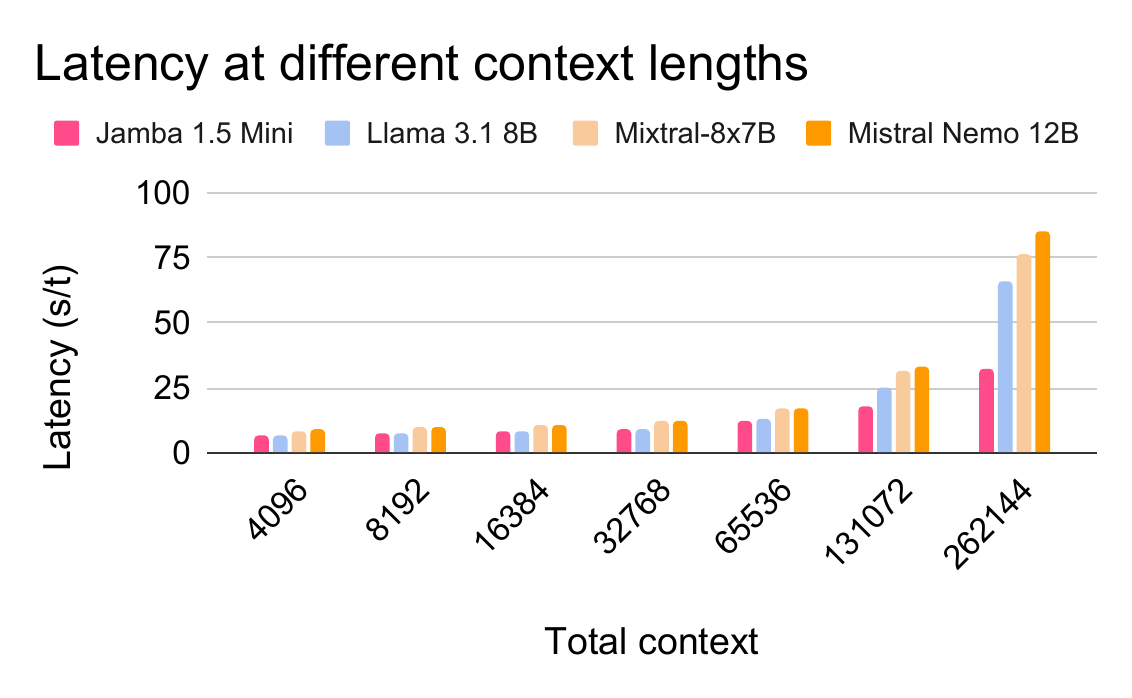}
    \caption{\jambamini, end-to-end latency.}
    \label{fig:jamba-mini-latency}
    \end{subfigure}
    \begin{subfigure}{.49\textwidth}
    \includegraphics[width=\linewidth]{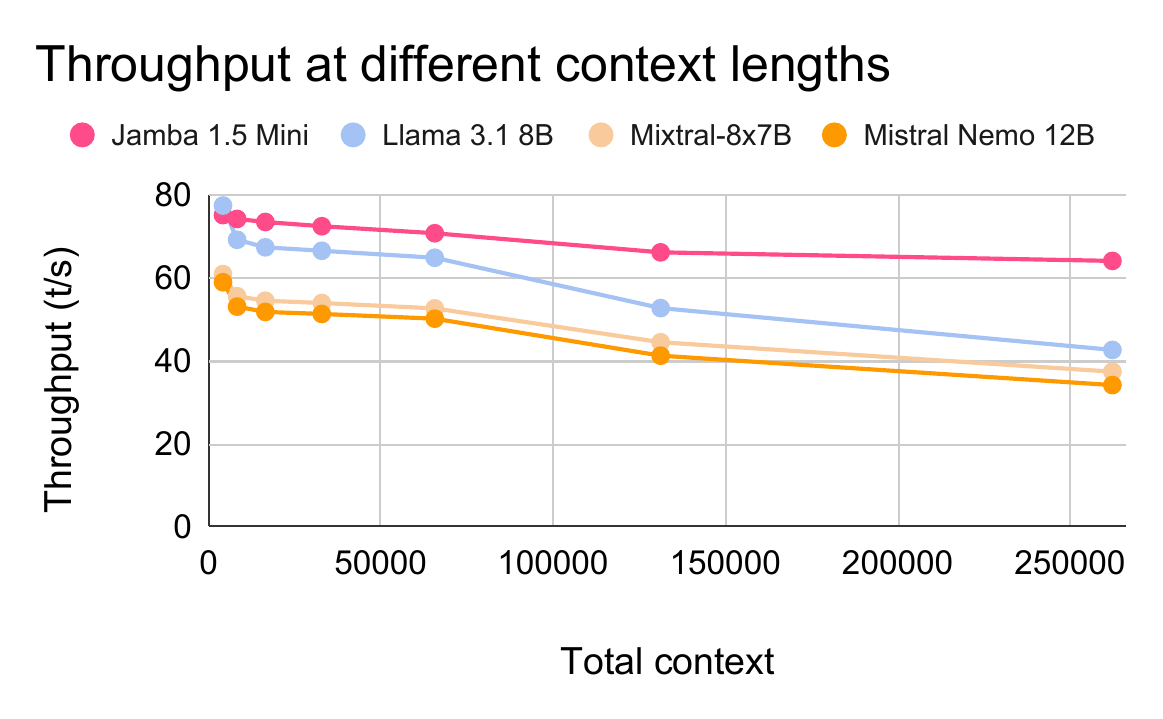}
    \caption{\jambamini, output tokens throughput.}
    \label{fig:jamba-mini-throughput}
    \end{subfigure}
    \caption{Comparison of \jambamini with other models in terms of latency and throughout. All measurements were done on 2xA100 80GB GPUs, with batch size 1, and output length 512 tokens.  \jambamini exhibits better latency, especially at large contexts, with only a slight reduction in output tokens throughput. }
    \label{fig:mini-throughput-latency}
    \vspace{-10pt}
\end{figure}

\begin{figure}[h]
    \centering
    \begin{subfigure}{.49\textwidth}
    \includegraphics[width=\linewidth]{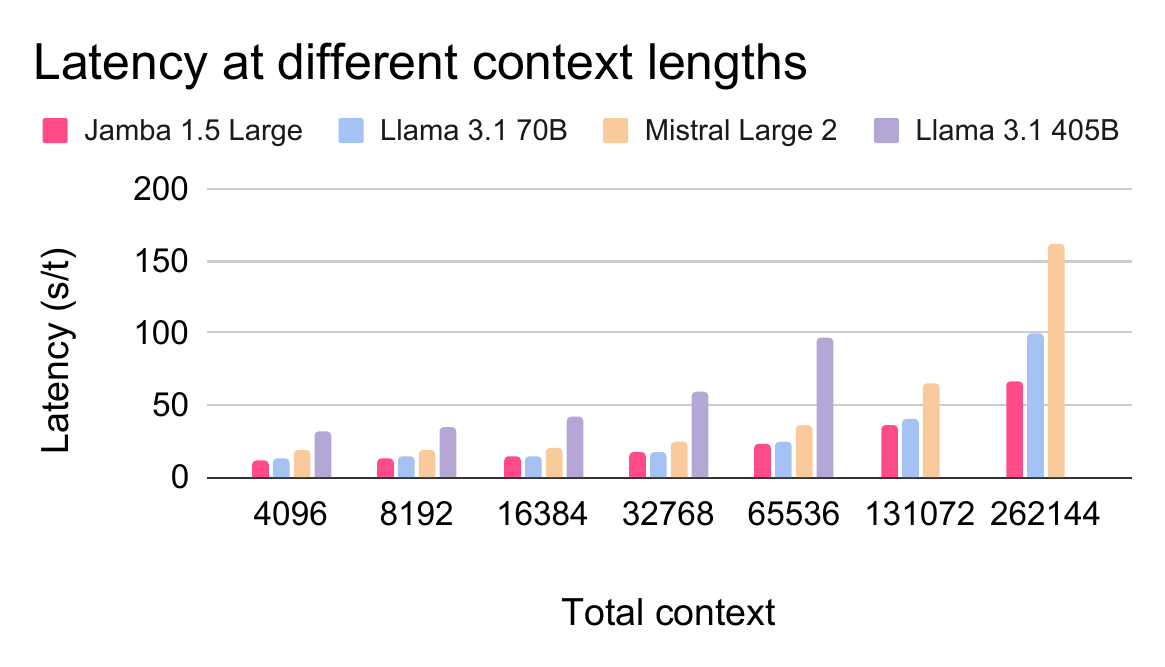}
    \caption{\jambalarge, end-to-end latency.}
    \label{fig:jamba-large-latency}
    \end{subfigure}
    \begin{subfigure}{.49\textwidth}
    \includegraphics[width=\linewidth]{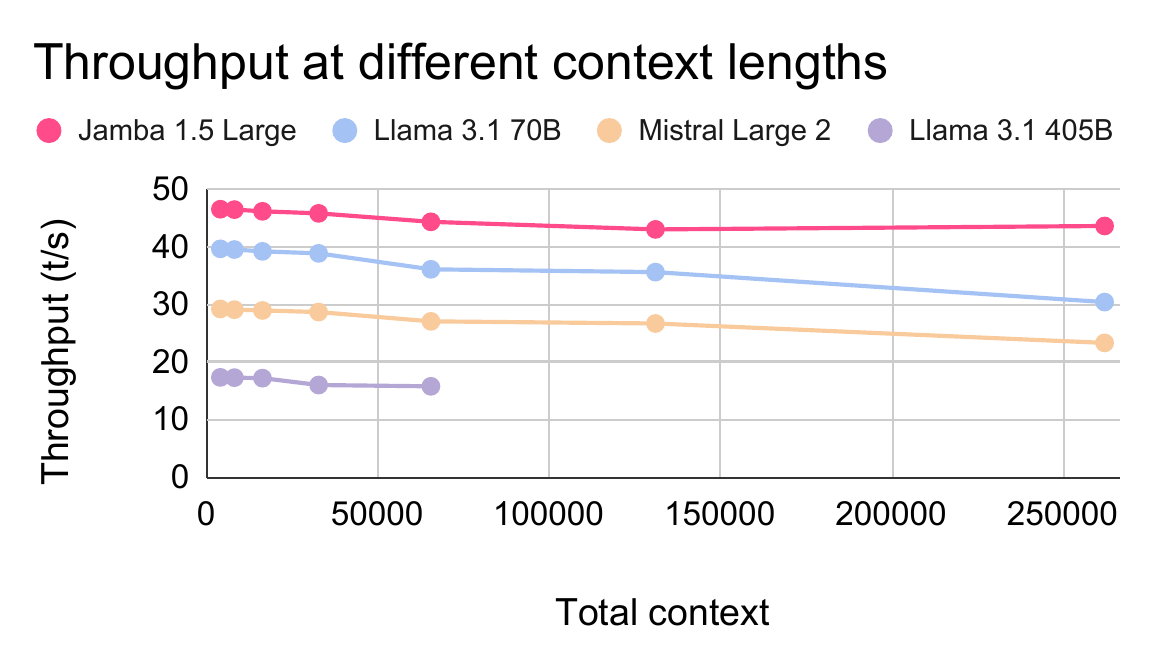}
    \caption{\jambalarge, output tokens throughput.}
    \label{fig:jamba-large-throughput}
    \end{subfigure}\\    
    \caption{Comparison of \jambalarge with other models in terms of latency and throughout. All measurements were done on 8xA100 80GB GPUs, with batch size 1, and output length 512 tokens. \jambalarge exhibits better latency, especially at large contexts, with only a slight reduction in output tokens throughput. The LLaMA-3.1-405B results truncate at 64K because the model is too large to fit context lengths greater than $\approx 100K$ tokens on 8 80GB GPUs. }
    \label{fig:large-throughput-latency}    
\end{figure}

Thanks to the hybrid \jamba architecture, our \jamba-1.5 models provide excellent throughput and latency. Figures \ref{fig:mini-throughput-latency} and \ref{fig:large-throughput-latency} show this for \jambamini and \jambalarge, respectively. 
As shown in the figures, our models obtain much better latency and throughput than similarly-sized models. Their advantage shines at long contexts, with substantial gaps. Importantly, \jambalarge runs efficiently even at long contexts, where the large LLaMA3-405B cannot run on the same hardware.

\vspace{-7pt}
\section{Training}
\vspace{-3pt}

\subsection{Training Infrastructure and Data}
\jambalarge  was trained on NVIDIA H100 GPUs using our in-house proprietary framework, which includes FSDP, tensor parallelism, sequence parallelism, and expert parallelism. For the latter we have adapted MegaBlocks \cite{megablocks}. 

\subsection{Training Stages}
The model was trained in three stages. During pre-training, it was first trained on an in-house dataset last updated in March 2024. Our pre-training dataset is a mixture of publicly available web documents, code, books and scientific articles. Our pre-processing pipeline  includes parsing, quality filters, and deduplication. To make the best use of publicly available data, we developed our own in-house parser, and used it to extract text and formatting. The exact data mixture was determined through various ablations. This stage included multilingual data with emphasis on the following languages: 
English, Spanish, French, Portueguse, Italian, Dutch, German, Arabic, and Hebrew.
It was then trained for a short phase of mid-training with a high proportion of long documents to emphasize its long-range capabilities. Finally, the model went through post-training, described in the next section. 

\subsection{Post-training} 
Our approach to post-training aims to achieve two objectives simultaneously: (\emph{i}) provide the model with various skills and conversational capabilities; (\emph{ii}) retain capabilities from pre-training and especially the long-context capabilities from mid-training. These two objectives are partly conflicting, since most of the available post-training datasets consist of relatively short examples. 

Given these considerations, our post-training process involves supervised fine-tuning \cite{weifinetuned,sanhmultitask} on high-quality conversational data, skill-specific data, and long-context data. 
Mixing these different types of data aims to retain long-context capabilities and acquire desired skills. 
As shown in the evaluations below, we find that our models perform very well in long-context evaluations. 

When performing supervised fine-tuning, we make heavy use of synthetic data, as is common in recent foundation models \cite{dubey2024LLaMA} and reflecting our approach for constructing structured data for building compound AI systems \cite{compoundAI2024}. We developed multiple different data synthesis pipelines, targeting different model capabilities. All pipelines apply the following pattern:
(i) Sample or generate prompts in a target distribution; (ii) 
Generate responses from language models; (iii)  
Filter or rank responses by quality according to automatic validation and scoring; and (iv)
Post-edit to remove artifacts and fit desired formatting. 
We use different models, prompting, sampling, filtering and editing for different data pipelines that compose the final data mixes.

We picked our final training recipes (data mix and hyperparameters) based on a battery of mostly internal automatic metrics.
Both \jamba-1.5 models are fine-tuned with the same control tokens and formatting template, which we provide as a part of our release as a HF-compatible tokenizer and chat template; see the model card for details.

We give several notable examples of synthetic data generation:
\paragraph{Table-based QA.} We generate tabular data and accompanying question-answer pairs, as demonstrated in our work on table understanding \cite{compoundAI2024}. We then convert the tables into natural language paragraphs using a language model. Our generated training examples include extraction, aggregation, and attribution tasks vis-a-vis text corresponding to specific rows or columns in a given table. 

\paragraph{Document QA.} Given a document, we prompt a language model to generate question-answer pairs, for both single and multiple paragraphs. We sometimes embed these examples within longer context by adding similar texts, to encourage long-context understanding with attribution. 

\paragraph{Tool use.}
We use the open-source Glaive function-calling dataset \cite{glaive} as a starting point, filtered with various heuristics and validations on the output schemas. To support parallel function calling, we first generate multiple valid parameter assignments for each function in Glaive. Next, we sample subsets of these valid parameter assignments, for the same function and across different functions, to generate user requests corresponding to the set of function calls. Finally, we prompt a function-calling language model to respond to these generated user requests and retaineonly responses where the function calls matched the original parameter assignments.

\paragraph{Steerability.}
We defined a set of instructions that can be easily validated and synthesized prompts that include a generic document-drafting task with one or more constraints added to it.
We generated completions for these prompts from a  language model and used rejection sampling based on the validations of our fine-grained instructions plus a general-purpose reward model.
To support instructions in system messages, we chose multiple prompts of this kind that share a fine-grained instruction instance and reformatted these prompts into a multi-turn conversation, with the instruction moved to the system message.

\subsection{Some Observations}

We share a few observations from the development of \jamba-1.5. While these are not fully explored, we hope they would inspire the community to look further into these issues.

First, while we included only a very small fraction of non-english data, for a few languages  and only for specific skills in the post-training phase, our \jamba-1.5 models perform quite well in multiple languages. We did include multilingual data in the pre-training phase, as mentioned above. Thus we speculate that the models are able to use the learned knowledge from that phase when being post-trained mostly in English. 

Second, our efficient \jamba architecture lowers the cost of fine-tuning on long contexts, allowing us to experiment more with a given budget. Thus we could experiment with multiple different training recipes at the post-training stage. 

Finally, while preference tuning algorithms like PPO \cite{schulman2017proximal} or DPO \cite{rafailov2024direct} improve alignment between  model outputs and human intent, we found that the combination of careful synthetic data generation, data filtering, and supervised fine-tuning is crucial for obtaining a strong post-trained model.

\section{Evaluation} \label{sec:evaluation}
\vspace{-4pt}

While we believe benchmarks are only partly correlated with success of real applications and user satisfaction, we report results on key public benchmarks. First, we report results on standard academic benchmarks. 
Then, we evaluate the model on chatbot benchmarks. Finally, we evaluate \jambalarge on several long-context evaluations and a multilingual evaluation. 
 
 We compare with recent open-weight models of the same size range: LLaMA-3.1 70B and Mistral-Large-2-123B when comparing with \jambalarge; LLaMA-3.1-8B and Gemma-2-9B when comparing with \jambamini.

\vspace{-3pt}
\subsection{Academic Benchmarks} \label{sec:evaluation-benchmarks}
\vspace{-1pt}

We report results with a wide range of  standard academic benchmarks: MMLU \cite{hendrycks2020measuring}, MMLU-Pro \cite{wang2024mmlu}, GPQA \cite{rein2023gpqa}, ARC-Challence  \cite{clark2018think}, BBH \cite{suzgun2023challenging}, and HumanEval \cite{chen2021evaluating}. 
We also evaluate on the IFEval instruction following dataset  \cite{zhou2023instruction} and the BFCL v1  function calling dataset \cite{berkeley-function-calling-leaderboard}.  Finally, we report safety evaluations on RealToxicity \cite{gehman2020realtoxicityprompts} and TruthfulQA \cite{lin2022truthfulqa}.

Table \ref{table:eval-benchmarks} compares \jambalarge to several publicly available models at similar sizes. 
All results are either taken from official sources or evaluated by us, as indicated in the table.\footnote{In two cases we failed to obtain good results: Mistral-Large-2 fails to obtain good scores on ARC-C despite multiple attempts. LLaMA-3.1 models perform poorly on GSM8K with the standard strict evaluation mode, so we also report for them a flexible evaluation, which allows higher results.}
We observe that the \jamba-1.5 models perform similarly to recent state-of-the-art publicly available models on standard academic benchmarks, including knowledge, reasoning, instruction following and function calling capabilities.  
We also observe similar safety metrics as those reported in the literature. We refer to Section \ref{sec:safety} for more information about our general approach for safety and alignment of models.

Importantly, the \jamba-1.5 models achieve these results while providing much better throughput and latency, as discussed above.

\begin{table}[h]
    \centering
    \resizebox{\textwidth}{!}{%
    \begin{tabular}{lc | cc >{\cellcolor{white}}c | cc >{\cellcolor{white}}c}
    \toprule    
 &  & \textcolor{ai21color}{\bf \jamba-1.5} & LLaMA-3.1 & Gemma-2 & \textcolor{ai21color}{\bf \jamba-1.5}   & LLaMA-3.1 & Mistral-L-2 \\ 
Benchmark & Metric & \textcolor{ai21color}{\bf Mini} & 8B & 9B & \textcolor{ai21color}{\bf Large}   & 70B & 123B \\ 
\midrule 
MMLU & 5-shot & 69.7 & 69.4 & 71.3 & 80.0   & 83.6 & \phantom{$^\dagger$}82.5$^\dagger$ \\ 
MMLU Pro & 5-shot & 39.8 & \phantom{$^\diamond$}38.0$^\diamond$ & \phantom{$^\diamond$}39.0$^\diamond$ & 48.3   & \phantom{$^\diamond$}53.0$^\diamond$ & \phantom{$^\dagger$}54.2$^\dagger$ \\ 
GPQA & 0-shot & 32.3 & \phantom{$^\diamond$}27.0$^\diamond$ & \phantom{$^\diamond$}36.0$^\diamond$ & 36.9   & \phantom{$^\diamond$}36.0$^\diamond$ & \phantom{$^\dagger$}40.7$^\dagger$ \\ 
ARC-C & 0-shot & 85.7 & 83.4 & 68.4 & 93.0   & 94.8 & \phantom{$^\dagger$}65.0$^\dagger$ \\ 
BBH & 3-shot & 53.4 & \phantom{$^\diamond$}51.0$^\diamond$ & \phantom{$^\diamond$}60.0$^\diamond$ & 65.5   & 69 & \phantom{$^\dagger$}70.8$^\dagger$ \\ 
HumanEval & pass@1 & 62.8 & 72.6 & 40.2 & 71.3   & 80.5 & 92 \\ 
GSM8K & 5-shot & 75.8 & \phantom{$^\star$}75.2/83.7$^\star$ & 68.6 & 87.0   & \phantom{$^\star$}71.5/94.2$^\star$ & \phantom{$^\dagger$}91.0$^\dagger$ \\ 
\midrule  
IFEval & 0-shot & 75.8 & 80.4 & 74.3 & 81.5   & 87.5 & \phantom{$^\dagger$}87.8$^\dagger$ \\ 
BFCL & 0-shot & 80.7 & 76.1 & \phantom{$^\ddagger$}-$^\ddagger$ & 85.5   & 84.8 & \phantom{$^\dagger$}85.1$^\dagger$ \\ 
\midrule 
RealToxicity & avg tox & 8.1 & - & 8.2 & 6.7   & - & - \\ 
TruthfulQA & 0-shot & 54.1 & \phantom{$^\dagger$}51.5$^\dagger$ & 50.2 & 58.3   & \phantom{$^\dagger$}60.7$^\dagger$ & \phantom{$^\dagger$}50.4$^\dagger$ \\    
    \bottomrule 
    \end{tabular}}
    \vspace{1pt}
    \caption{\jamba-1.5 models obtain similar performance to similarly sized models while enjoying a better throughput and latency. $^\dagger$ evaluation run by us. $^\diamond$ reported in the HuggingFace OpenLLM leaderboard. $^\ddagger$ Lacking function calling capabilities. $^\star$ Strict/flexible evaluation.}
    \label{table:eval-benchmarks}
    \vspace{-15pt}
\end{table}

\vspace{-3pt}
\subsection{ChatBot Evaluations}

In this section we evaluate the \jamba-1.5 models on two chatbot scenarios: 
Arena-Hard \cite{arenahard2024}, a set of 500 challenging user queries that uses GPT4-Turbo as a judge, and WildBench \cite{Lin2024WildBenchBL}, which uses GPT4-Turbo as a judge with a length bias mitigation. 
As Table \ref{table:eval-chatbot} shows, \jamba-1.5 models obtain excellent reuslts in these evaluations, with \jambalarge surpassing LLaMA-3.1 70B, but somewhat trailing behind Mistral-Large-2 123B, which has about 30\% more active parameters.

\begin{table}[h!]
\vspace{-5pt}
    \centering
    \resizebox{\textwidth}{!}{%
    \begin{tabular}{l | cc >{\cellcolor{white}}c | cc >{\cellcolor{white}}c}
    \toprule 
              &    \textcolor{ai21color}{\bf \jamba-1.5}   &    LLaMA-3.1 & Gemma-2 & \textcolor{ai21color}{\bf \jamba-1.5} &  LLaMA-3.1 &   Mistral-L-2  \\  
    Benchmark  &  \textcolor{ai21color}{\bf Mini}  &   8B    &    9B & \textcolor{ai21color}{\bf Large}   &    70B & 123B  \\ 
    \midrule 
    Arena-Hard & 46.1 & 21.3 & \phantom{$^\dagger$}43.2$^\dagger$ & 65.4 & 55.7 & 70.4  \\ 
    Wild-Bench & 42.4 & \phantom{$^\dagger$}33.6$^\dagger$ & 42.7  & 48.5 & \phantom{$^\dagger$}49.8$^\dagger$ & \phantom{$^\dagger$}56.3$^\dagger$ \\ 
    \bottomrule
    \end{tabular}}
    \vspace{1pt}
    \caption{Comparison of \jamba-1.5 models to similarly sized models on chatbot benchmarks.  \jamba-1.5 models obtain similar performance with better throughput and latency. $^\dagger$ evaluation run by us.}
    \label{table:eval-chatbot}
    \vspace{-10pt}
\end{table}


\subsection{Long-Context Evaluations} \label{sec:evaluation-long}
\vspace{-3pt}

The released model handles context lengths of up to 256K tokens. In this section, we evaluate it on synthetic and naturalistic benchmarks that test its long-context capabilities. 

\subsubsection{RULER}
We evaluate on the RULER benchmark, a set of 13 synthetic tasks aimed to assess long-context capabilities of language models. RULER includes 8 variants of needle-in-a-haystack retrieval tasks \cite{kamradt2023,mohtashami2023randomaccess,li2023long,Liu2023LostIT}, including multiple `needles' \cite{arorazoology}.  It also has one variable tracking task where a chain of variable bindings should be returned, two aggregation tasks where one needs to return the most common words, and two question-answering tasks, where paragraphs cotraining answers from naturalistic datasets \cite{rajpurkar2018know,Yang2018HotpotQAAD} are inserted into random paragraphs to simulate long contexts. 

The results are shown in Table \ref{table:ruler}. 
Among all publicly available and proprietary models, \jambamini and \jambalarge are the only ones with a confirmed effective length of 256K tokens. Gemini-pro reports good results up to 128K on the original RULER paper. However, we were unable to reproduce these results despite much effort. We examined Gemini-pro generations and noticed the model often fails to answer or generates a refusal. Since the official RULER results are from a preview version, we hypothesize that Gemini-pro had since undergone through updates that have hurt its performacne on RULER.

\begin{table}[h]
    \centering
    \resizebox{\textwidth}{!}{%
    \begin{tabular}{l | cc | c c c c c c c | c }
        \toprule 
        & Claimed & Effective & & & & & & & &  \\
        & Length & Length   & 4k & 8k & 16k & 32k & 64k & 128k & 256k & Avg.  \\ 
        \midrule 
\textcolor{ai21color}{\bf \jambalarge} & 256K & \textbf{256K} & \underline{96.7} & \underline{96.6} & \underline{96.4} & \underline{96.0} & \underline{95.4} & \underline{95.1} & \underline{93.9} & \underline{95.7} \\ 
\textcolor{ai21color}{\bf \jambamini} & 256K & \textbf{256K} & \underline{95.7} & \underline{95.2} & \underline{94.7} & \underline{93.8} & \underline{92.7} & \underline{89.8} & \underline{86.1} & 92.6 \\ 
\midrule 
Gemini-1.5-pro & 1M & >128K & \underline{96.7} & \underline{95.8} & \underline{96} & \underline{95.9} & \underline{95.9} & \underline{94.4} & \phantom{$^\dagger$}65.1$^\dagger$ & 91.4 \\ 
GPT-4-1106-preview & 128K & 64K & \underline{96.6} & \underline{96.3} & \underline{95.2} & \underline{93.2} & \underline{87} & 81.2 & - & 91.6 \\
LLaMA 3.1 70B & 128K & 64K & \underline{96.5} & \underline{95.8} & \underline{95.4} & \underline{94.8} & \underline{88.4} & 66.6 & - & 89.6 \\
Qwen2 72B & 128K & 32K & \underline{96.9} & \underline{96.1} & \underline{94.9} & \underline{94.1} & 79.8 & 53.7 & - & 85.9 \\
Command-R+ & 128K & 32K & \underline{95.6} & \underline{95.2} & \underline{94.2} & \underline{92} & 84.3 & 63.1 & - & 87.4 \\
LLaMA 3.1 8B & 128K & 32K & \underline{95.5} & \underline{93.8} & \underline{91.6} & \underline{87.4} & 84.7 & 77 & - & 88.3 \\
Command-R & 128K & 32K & \underline{93.8} & \underline{93.3} & \underline{92.4} & \underline{89.5} & 84.9 & 76 & - & 88.3 \\
Mistral Large 2 & 128K & 32K & \underline{96.2} & \underline{96.1} & \underline{95.1} & \underline{93} & 78.8 & 23.7 & - & 80.5 \\
Mixtral 8x22B & 64K & 32K & \underline{95.6} & \underline{94.9} & \underline{93.4} & \underline{90.9} & 84.7 & 31.7 & - & 81.9 \\
Yi 34B & 200K & 32K & \underline{93.3} & \underline{92.2} & \underline{91.3} & \underline{87.5} & 83.2 & 77.3 & - & 87.5 \\
Phi3 mini 3.8B & 128K & 32K & \underline{92.2} & \underline{91.5} & \underline{90.7} & \underline{87.5} & 80.6 & 66.7 & - & 84.8 \\
Phi3 medium 14B & 128K & 32K & \underline{93.3} & \underline{93.2} & \underline{91.1} & \underline{86.8} & 78.6 & 46.1 & - & 81.5 \\
Mixtral 8x7B & 32K & 32K & \underline{94.9} & \underline{92.1} & \underline{92.5} & \underline{85.9} & 72.4 & 44.5 & - & 80.4 \\
Mistral Nemo 12B & 128K & 16K & \underline{87.8} & \underline{87.2} & \underline{87.7} & 69 & 46.8 & 19 & - & 66.2 \\
DBRX & 32K & 8K & \underline{95.1} & \underline{93.8} & 83.6 & 63.1 & 2.4 & 0 & - & 56.3 \\     
\bottomrule 
    \end{tabular}}
    \vspace{1pt}
    \caption{Comparison of \jamba-1.5 models with other publicly available and proprietary models on the RULER benchmark.  Results for other models are from the RULER Github. $^\dagger$ Evaluation run by us. \jamba-1.5 models are the only ones with a confirmed effective length of 256K tokens. }
    \label{table:ruler}
    \vspace{-15pt}
\end{table}

\subsubsection{Infinite-\textsc{Bench}} 
Next we evaluate on  $\infty$\textsc{Bench}, a dataset designed to evaluate long-context abilities of language models, with an average length of 100K tokens.  We focus on two English tasks on understanding long novels: question answering (EN.QA) and multiple-choice question answering (EN.MC). 
As Table~\ref{table:eval-infty} shows, \jamba-1.5 models perform very well in this case, outperforming similarly sized LLaMA-3.1 and Mistral-Large-2 models. (We do not report results with Gemma-2 9B due to its short context window of 8K.)

\begin{table}[h!]
\vspace{-3pt}
    \centering
    \resizebox{\textwidth}{!}{%
    \begin{tabular}{l | cc >{\cellcolor{white}}c | cc >{\cellcolor{white}}c}
    \toprule 
              &    \textcolor{ai21color}{\bf \jamba-1.5}   &    LLaMA-3.1 & Gemma-2 & \textcolor{ai21color}{\bf \jamba-1.5} &  LLaMA-3.1 &   Mistral-L-2  \\  
    Benchmark  &  \textcolor{ai21color}{\bf Mini}  &   8B    &    9B & \textcolor{ai21color}{\bf Large}   &    70B & 123B  \\ 
    \midrule 
    EN.MC & 76.9 & 65.1 & -  & 80.4 &  78.2 & \phantom{$^\dagger$}36.9$^\dagger$  \\  
    EN.QA & 40.6 & 27.1 & -  & 34.9 &  36.7 & -  \\
    \bottomrule 
    \end{tabular}}
    \vspace{1pt}
    \caption{\jamba-1.5 models outperform similarly sized LLaMA-3 and Mistral-Large-2 models in long-context evaluations. $^\dagger$ evaluation run by us.     }
    \label{table:eval-infty}
    \vspace{-3pt}
\end{table}

\subsection{Multilingual capabilities}

We perform a basic evaluation of \jamba-1.5 abilities in non-English langauges. In particular, we report results on the multilingual MMLU dataset \cite{lai2023okapi} as distributed through the LM Evaluation Harness \cite{eval-harness}. 
Table~\ref{table:eval-multilingual} shows the results, where \jambamini performs similarly or better than its points of comparison. \jambalarge is slightly behind its comparable models, but still exhibits good multilingual capabilities.

\begin{table}[h]
    \centering
    \resizebox{\textwidth}{!}{%
    \begin{tabular}{l ccccccc | c}
    \toprule 
  &  Spanish  &  Portuguese  &  French  &  German  &  Arabic  &  Italian  &  Dutch  &  Avg \\ 
  \midrule 
\textcolor{ai21color}{\bf \jambamini}  &  66.3  &  66.7  &  65.9  &  63.8  &  57.3  &  65.1  &  65.0  &  64.30 \\   
LLaMA-3.1-8B  &  59.5  &  59.1  &  59.5  &  57.2  &  46.9  &  58.4  &  57.2  &  56.83 \\ 
Gemma-9B  &  66.0  &  59.9  &  66.7  &  64.3  &  55.9  &  65.8  &  64.8  &  63.34 \\ 
\midrule 
\textcolor{ai21color}{\bf \jambalarge }  &  75.5  &  75.5  &  75.8  &  73.9  &  67.1  &  75.2  &  74.6  &  73.94 \\ 
LLaMA-3.1-70B  &  79.5  &  79.4  &  79.1  &  78.4  &  70.4  &  79.1  &  78.4  &  77.76 \\ 
Mistral-Large-2  &  78.7  &  78.4  &  78.4  &  77.4  &  65.9  &  78.3  &  76.2  &  76.19 \\ 
\bottomrule 
    \end{tabular}}
    \vspace{1pt}
    \caption{Comparison of \jamba-1.5 with other models on the multilingual MMLU dataset. }
    \label{table:eval-multilingual}
\end{table}

\vspace{-5pt}
\section{Alignment and Safety Considerations} \label{sec:safety}

Our approach to alignment of our models is driven by creating transparency between model behavior and customer expectations. Our models default to a business code of conduct based on our participation in industry standards bodies, think tanks and direct experience with our customers. We see this as an ongoing and evolving collaboration. In addition, companies have multiple ways to control model conduct to reflect their individual values and cultures such as additional training and fine tuning, system messages and prompt engineering. Overall, our AI code of conduct is based on the following objectives:

\begin{itemize}[itemsep=2pt,topsep=2pt,parsep=2pt]
\item Align model behavior and output with company values and normative business decorum.
\item Clearly state tenets of intended behavior such that errors/bugs are easily discerned.
\item Collaborate with Customers and map behavior to their best practices.
\item Continuously gather feedback to monitor and actively improve behavior.
\end{itemize}

In line with our role in an OECD task force to develop a monitoring mechanism for applying the G7 Hiroshima Code of Conduct for Organisations Developing Advanced AI Systems, we have organized our model alignment work with the OECD values-based AI principles:\footnote{\url{https://oecd.ai/en/ai-principles}} inclusive growth, sustainable development and well-being; 
human-centered values and fairness;
transparency and explainability;
robustness, security and safety; and 
accountability. 

For each of the first four principles we have detailed behavioral expectations or tenets and examples that can be used to train/align and test for compliance. The principle of accountability is focused on AI21’s role in taking responsibility for the behavior of the models. We submit that this accountability is demonstrated primarily through transparency and engagement with customers, regulators and independent 3rd-parties. Our engagement with OECD, Stanford’s HELM \cite{Liang2023HolisticEO} and FMTI \cite{Bommasani2024TheFM} and documents like this demonstrate this commitment, as well as our high ranking on the FMTI (2nd as of May 2024).  

In total, we have created 60 tenets that map to the OECD principles. These tenets are stated as directives of behavior for our models to avoid.  The full list will be made publicly available.

\section{Conclusion}

We have presented \jambalarge and \jambamini, two new large-scale models based on the \jamba hybrid Transformer-Mamba architecture. Both models achieve excellent performance in academic benchmarks, chatbot evaluations, and long-context evaluations, while offering improved latency and throughput, especially for long contexts. We release the model weights for use by the community in hopes that others build on this technology. 

\clearpage 

\section*{Contributions}

\renewcommand*{\thefootnote}{*}

\begin{multicols*}{2}
\textbf{Pre- and Post-Training} \\ 
Alan Arazi \\ 
Barak Lenz{\hypersetup{hidelinks}\footnote{Project leads}} \\ 
Chen Almagor \\ 
Dan Padnos\textsuperscript{*} \\ 
Daniel Gissin\textsuperscript{*} \\ 
Daniel Jannai \\ 
Dor Muhlgay \\ 
Edden M Gerber \\ 
Erez Safahi \\ 
Gal Cohen \\ 
Gal Shachaf \\ 
Hofit Bata \\ 
Inbal Magar \\ 
Itay Dalmedigos \\ 
Jhonathan Osin\textsuperscript{*} \\ 
Matan Danos \\ 
Michael Gokhman \\ 
Nir Ratner \\ 
Noam Gat \\ 
Noam Rozen \\ 
Omer Antverg \\ 
Omri Abend \\ 
Opher Lieber\textsuperscript{*} \\ 
Orit Cohavi \\ 
Raz Alon \\ 
Shaked Meirom \\ 
Tom Braude \\ 
Uriya Pumerantz \\ 
Yonatan Belinkov \\ 
Yuval Globerson \\ 
Yuval Peleg Levy \\

\textbf{Serving \& Infrastructure} \\ 
Amir Bergman \\ 
Avshalom Manevich \\ 
Barak Peleg \\ 
Elad Dolev \\ 
Eran Krakovsky \\ 
Erez Schwartz \\ 
Haim Rozenblum \\ 
Mor Zusman \\ 
Oded Fried \\ 
Roman Glozman \\ 
Shahar Lev \\ 
Tomer Asida \\ 
Yehoshua Cohen \\ 

 \columnbreak
 
\textbf{Data} \\ 
Ben Aviram \\ 
Dor Zimberg \\ 
Ido Blass \\ 
Ohad Leshno \\ 
Rom Gilad \\ 
Tom Ben Gal \\ 

\textbf{Evaluation} \\ 
Clara Fridman \\ 
Julie Fadlon \\ 
Maria Rozman \\ 
Naama Gidron \\ 
Ro'i Belson \\ 
Tal Ness \\ 

\textbf{Project \& Product Management} \\ 
Or Dagan\textsuperscript{*} \\ 
Roi Cohen \\ 
Shaked Meirom\textsuperscript{*} \\ 
Tal Delbari \\ 
Yoav Shoham \\

\end{multicols*}

\clearpage 

\bibliographystyle{plain}
\bibliography{refs}

\end{document}